\documentclass[final]{cvpr}

\usepackage{times}
\usepackage{epsfig}
\usepackage{graphicx}
\usepackage{amsmath}
\usepackage{amssymb}
\usepackage{color,soul}
\usepackage{subcaption}
\usepackage{siunitx}
\usepackage{makecell}
\usepackage{nopageno}
\usepackage[pagebackref=true,breaklinks=true,colorlinks,bookmarks=false]{hyperref}

\def\cvprPaperID{15} 
\def\confYear{CVPR 2021}

\title{DeepDarts: Modeling Keypoints as Objects for\\Automatic Scorekeeping in Darts using a Single Camera}

\author{William McNally\quad Pascale Walters\quad Kanav Vats\quad Alexander Wong\quad John McPhee\\
Systems Design Engineering, University of Waterloo, Canada\\
Waterloo Artificial Intelligence Institute, University of Waterloo, Canada\\
{\tt\small \{wmcnally, pbwalter, k2vats, a28wong, mcphee\}@uwaterloo.ca}
}

\begin{document}

\twocolumn[\maketitle\vspace{-3em}\begin{center}
    \includegraphics[width=1.0\linewidth]{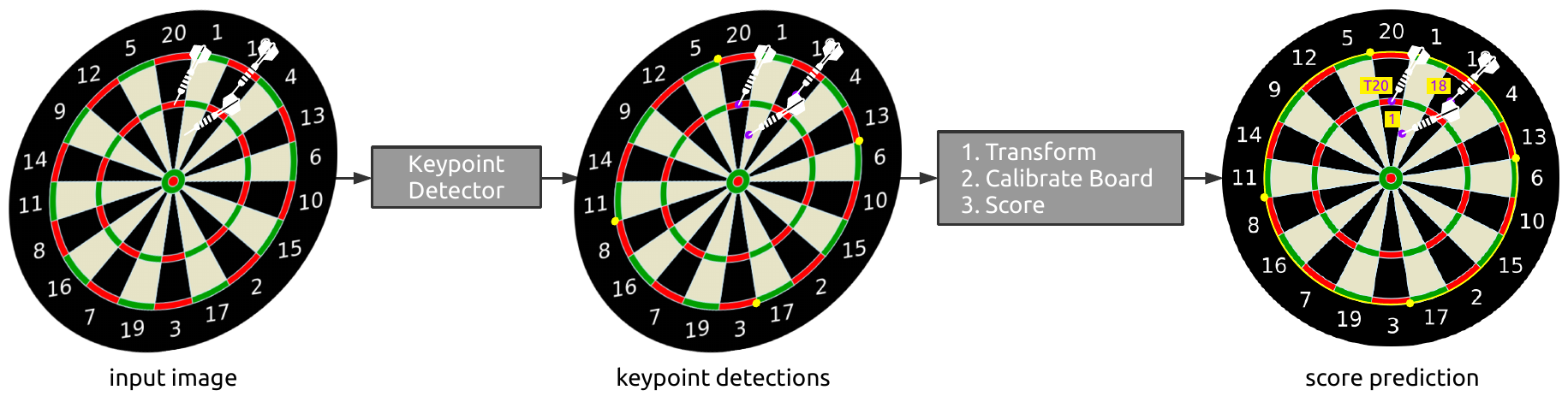}
\end{center}
\vspace{-12pt}
\captionof{figure}{Overview of DeepDarts, a method for predicting dart scores from a single image captured from any camera angle. DeepDarts uses a new deep learning-based keypoint detector that models keypoints as objects to simultaneously detect and localize the dart coordinates (purple) and four dartboard calibration points (yellow). The calibration points are used to transform the dart locations to the dartboard plane and calibrate the scoring area. The dart scores are then classified based on their position relative to the center of the dartboard.}
\vspace{-4pt}
\label{fig:teaser}
\bigbreak]

\begin{abstract}
\vspace{-10pt}
Existing multi-camera solutions for automatic scorekeeping in steel-tip darts are very expensive and thus inaccessible to most players. Motivated to develop a more accessible low-cost solution, we present a new approach to keypoint detection and apply it to predict dart scores from a single image taken from any camera angle. This problem involves detecting multiple keypoints that may be of the same class and positioned in close proximity to one another. The widely adopted framework for regressing keypoints using heatmaps is not well-suited for this task. To address this issue, we instead propose to model keypoints as objects. We develop a deep convolutional neural network around this idea and use it to predict dart locations and dartboard calibration points within an overall pipeline for automatic dart scoring, which we call DeepDarts. Additionally, we propose several task-specific data augmentation strategies to improve the generalization of our method. As a proof of concept, two datasets comprising 16k images originating from two different dartboard setups were manually collected and annotated to evaluate the system. In the primary dataset containing 15k images captured from a face-on view of the dartboard using a smartphone, DeepDarts predicted the total score correctly in 94.7\% of the test images. In a second more challenging dataset containing limited training data (830 images) and various camera angles, we utilize transfer learning and extensive data augmentation to achieve a test accuracy of 84.0\%. Because DeepDarts relies only on single images, it has the potential to be deployed on edge devices, giving anyone with a smartphone access to an automatic dart scoring system for steel-tip darts. The code and datasets are available\footnote{ \href{https://github.com/wmcnally/deep-darts}{https://github.com/wmcnally/deep-darts}.}.
\end{abstract}
\vspace{-10pt}

\section{Introduction}
\vspace{-4pt}
Deep learning-based computer vision has recently gained traction in the sports industry due to its ability to autonomously extract data from sports video feeds that would otherwise be too tedious or expensive to collect manually. Example sports applications include field localization~\cite{homayounfar2017sports}, player detection and tracking~\cite{thaler2013real, cioppa2020multimodal, ullah2018directed}, equipment and object tracking~\cite{voeikov2020ttnet, vats2019pucknet, reno2018convolutional}, pose estimation~\cite{bridgeman2019multi, nekoui2020falcons, arbues2020using}, event detection~\cite{mcnally2019golfdb, giancola2018soccernet, cai2019temporal, vats2020event, sanford2020group}, and scorekeeping~\cite{voeikov2020ttnet}. This data is often more informative than conventional human-recorded statistics and as such, the technology is quickly ushering in a new era of sports analytics. In this work, we explore the use of deep learning-based computer vision to perform automatic scorekeeping in steel-tip darts. 

While darts is a sport that is played professionally under multiple governing bodies, it is better known as the traditional pub game that is played recreationally around the world. Typically, it is the responsibility of the player to keep their own score, and doing so requires quick mental math. In ``501,'' the most widely played game format, the player must add up the individual scores of each dart and subtract this amount from their previous total. As trivial as this may sound, scorekeeping in darts slows down the pace of the game and arguably makes it less enjoyable. 

Several automated scoring systems have therefore been proposed to improve the playability of darts. A prerequisite for these systems is the precise location of dart landing positions relative to the dartboard, so these systems can additionally provide statistics based on dart positional data. Electronic dartboards have been used together with plastic-tip darts to enable automatic scoring. However, this variation of the sport known as soft-tip darts lacks the authenticity of traditional steel-tip darts played on a bristle dartboard, and is much less popular as a result. For this reason, multi-camera systems have been developed for automatic scoring in steel-tip darts\footnote{Examples of such systems are sold under the brand names Scolia, Spiderbull, Dartsee, and Prodigy.}. These systems position cameras around the circumference of the dartboard and use image processing algorithms to locate the darts on the board. While these multi-camera systems are accurate and subtle, they are expensive and only function with the dartboard for which they were designed. Therefore, they are not very accessible. Moreover, the cameras are susceptible to damage from wayward darts due to their close proximity to the dartboard.

In an effort to develop a more accessible automated scoring system for steel-tip darts, we investigate the feasibility of using deep learning to predict dart scores from a single image taken from any front-view camera angle. To the best of our knowledge, no like software currently exists. We propose a new approach to keypoint detection that involves modeling keypoints as objects, which enables us to simultaneously detect and localize multiple keypoints that are of the same type and clustered tightly together. Regressing keypoints using heatmaps, which is currently the \textit{de facto} standard for keypoint estimation~\cite{mcnally2020evopose2d, cheng2020higherhrnet, sun2019deep, xiao2018simple, chen2018cascaded, cao2017realtime, newell2016stacked, huang2020awr, iqbal2018hand, dong2018style}, does not adequately address this problem. We develop a deep convolutional neural network (CNN) around this idea and use it to detect four dartboard calibration points in addition to the dart landing positions. The predicted calibration points are used to map the predicted dart locations to a circular dartboard and calibrate the scoring area. The dart scores are then classified based on their relative position to the center of the dartboard. We refer to this system as DeepDarts (see Fig.\ \ref{fig:teaser}). 

Our research contributions are summarized as follows: \textbf{(i)} We develop a new deep learning-based solution to keypoint detection and apply it to predict dart scores from a single image taken from any camera angle. \textbf{(ii)} We contribute two dartboard image datasets containing a total of 16k dartboard images and corresponding labels for the dart landing positions and four dartboard calibration points. We additionally propose a task-specific evaluation metric that takes into account false positives and negatives and is easy to interpret. \textbf{(iii)} Finally, we propose several task-specific data augmentation strategies and empirically demonstrate their generalization benefits. 

\section{Related Work}
\vspace{-4pt}
To the best of our knowledge, there are no published works in the computer vision literature that focus on the problem of predicting dart scores from images. Perhaps the most closely related works are those using non-deep learning image processing for automatic scoring in range shooting~\cite{ali2008computer, ding2009design, aryan2012vision} and archery~\cite{zin2013image, Parag2017SequentialRA}. These methods use traditional image processing algorithms to engineer features for detection and scoring. However, deep learning-based detection methods offer superior performance as they learn robust high-level features directly from the data. Furthermore, deep learning-based methods are better equipped to handle occlusion, variations in viewpoint, and illumination changes~\cite{zhao2019object}. Our method for predicting dart scores from single images was inspired by the deep learning literature related to keypoint and object detection, and so we discuss these research areas in the following sections.

\smallskip\noindent\textbf{Keypoint detection} involves simultaneously detecting objects and localizing keypoints associated with those objects. A common application of keypoint detection is 2D human pose estimation~\cite{toshev2014deeppose}, which involves localizing a set of keypoints coinciding with various anatomical joints for each person in an image. The two-stage ``top-down'' approach is the most common~\cite{mcnally2020evopose2d, sun2019deep}, where an off-the-shelf object detection CNN is first used to find the people in the image and then a second CNN localizes the keypoints for each person instance. The second network learns the keypoint locations by minimizing the mean squared error between predicted and target \textit{heatmaps}~\cite{tompson2014joint}, where the latter contain 2D Gaussians centered on the ground-truth keypoint locations. A separate heatmap is predicted for each keypoint type, or class. The heatmap regression method is also used in related keypoint estimation tasks, including hand pose estimation~\cite{iqbal2018hand} and facial landmark detection~\cite{dong2018style}.

Running two CNNs in series is not cost-efficient, so alternative human pose estimation methods have been proposed that bypass the initial person detection stage. These are referred to as ``bottom-up'' approaches because they first localize all the keypoints in the image and then assign them to different person instances~\cite{cao2017realtime, cheng2020higherhrnet}. Bottom-up approaches save computation, but they are generally less accurate. This is in part due to the scale of the people in the image with respect to the spatial resolution of the heatmaps, but also due to the fact that when two keypoints of the same class appear very close together, their heatmap signals overlap and can be difficult to isolate (e.g., a right knee occludes another right knee, see Fig.\ 9c in~\cite{cao2017realtime}). If the bottom-up approach were applied to the problem of predicting dart locations, overlapping heatmap signals would be an issue as darts are often clustered tightly together. On the other hand, the top-down keypoint detection pipeline demands excessive computation that does not favor deployment on edge devices. Importantly, our proposed approach to keypoint detection adequately addresses both of these issues. 

\smallskip\noindent\textbf{Object detection} involves detecting instances of objects of a certain class within an image, where each object is localized using a rectangular bounding box. Similar to other image recognition tasks, deep learning approaches exploiting CNNs have demonstrated proficiency in object detection. These methods can be categorized into two main types based on whether they use one or two stages. The two-stage approach is reminiscent of a more traditional object detection pipeline, where region proposals are generated and then classified into different object categories. These methods include the family of R-CNN models~\cite{girshick2014rich, girshick2015fast, ren2015faster, he2017mask, cai2018cascade}. Single-stage object detection models regress and classify bounding boxes directly from the image in a unified architecture. Examples of these methods include the Single Shot MultiBox Detector (SSD)~\cite{liu2016ssd}, RetinaNet~\cite{lin2017focal}, and the ``You Only Look Once'' (YOLO) family of models~\cite{redmon2016you, redmon2017yolo9000, redmon2018yolov3, bochkovskiy2020yolov4}. Single-stage object detectors offer superior computational efficiency, and are able detect objects in real-time~\cite{redmon2016you}. Moreover, it has recently been shown that single-stage object detectors can be scaled to achieve state-of-the-art accuracy while maintaining an optimal accuracy-speed trade-off~\cite{wang2020scaled}. 

Because object detectors are often used as a preliminary to keypoint detection, there is significant overlap between the two streams of research. Mask R-CNN was used to predict keypoints by modeling keypoint locations using a one-hot mask~\cite{he2017mask}. However, the accuracy of this approach is inherently limited by the spatial resolution of the one-hot mask. Conversely, keypoint estimators have been repurposed for object detection. Zhou et al.\ proposed to model objects as points by regressing object centers using heatmaps~\cite{zhou2019objects}. Our approach to keypoint detection can be viewed as the opposite to that of Zhou et al., where we instead model keypoints as objects to mitigate the drawbacks of heatmaps when multiple keypoints of the same type exist in an image. 

\section{Dart Scoring and Terminology}
\begin{figure}
\vspace{-12pt}
\centering
    \includegraphics[trim={0, 2mm, 0, 0}, clip, width=1\linewidth]{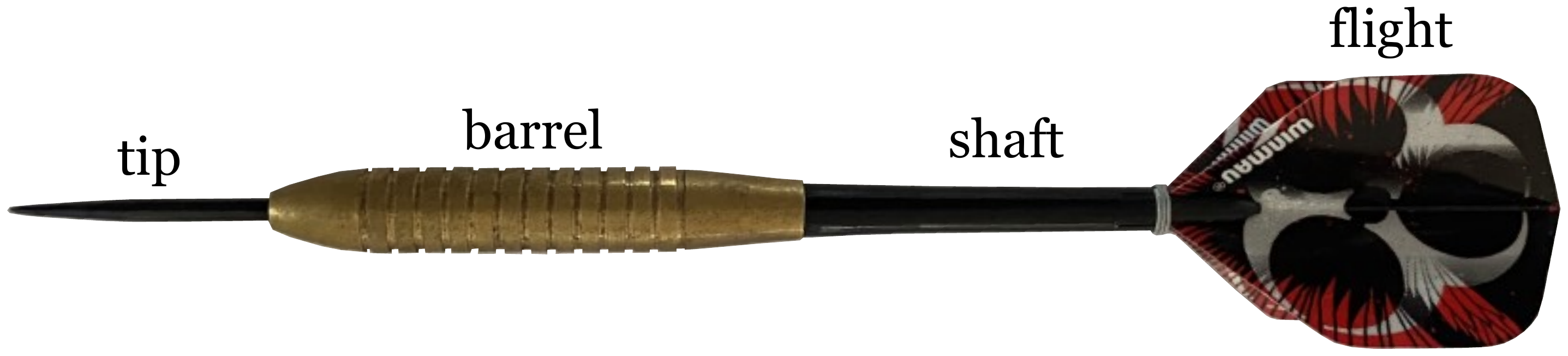}
\vspace{-20pt}
\caption{The four components of a steel-tip dart.}
\vspace{-16pt}
\label{fig:dart}
\end{figure}

\vspace{-4pt}
Darts is a sport in which pointed projectiles (the darts) are thrown at a circular target known as a dartboard. A dart is made of four components, including the tip, barrel, shaft, and flight. These components are indicated in Fig.\ \ref{fig:dart}. Points are scored by hitting specific marked areas of the dartboard. The modern dartboard is divided into 20 numbered sections scoring 1 to 20 points (see Fig.\ \ref{fig:teaser}). Two small circles are located at the center of the dartboard; they are known collectively as the bullseye. The inner red circle of the bullseye is commonly referred to as ``double bull'' (DB) and is worth 50 points, whereas the outer green circle is typically referred to simply as ``bull'' (B) and is worth 25 points. The ``double ring'' is the thin red/green outer ring and scores double the points value of that section. The ``treble ring'' is the thin red/green inner ring and scores triple the points value of that section. Typically, three darts are thrown per turn, so the maximum attainable score for a single turn is 180, by scoring three triple-20s (T20).

\section{DeepDarts}
\vspace{-4pt}
DeepDarts is a system for predicting dart scores from a single image taken from any camera angle. It consists of two stages: \textit{keypoint detection} and \textit{score prediction}. DeepDarts takes on a new approach to keypoint detection, in which keypoints are modeled as objects. We discuss each stage of the system in more detail in the following sections. Finally, several data augmentation strategies are proposed to improve the accuracy of the dart score predictions.

\subsection{Modeling Keypoints as Objects for Dartboard Keypoint Detection}
\vspace{-4pt}
Predicting dart scores demands a system that can precisely locate the exact coordinates where the dart tips strike the dartboard. While this problem shares similarities with 2D keypoint regression (e.g., 2D human pose estimation, hand pose estimation, and facial landmark detection), there are two key differences: (i) the total number of keypoints is not known \textit{a priori}, as there may be any number of darts present in a given dartboard image, and (ii) the darts are indistinguishable from one another and thus cannot be assigned to different keypoint classes. The widely adopted framework for regressing 2D keypoints using heatmaps is ill-equipped to handle this task because when multiple darts appear close together, their heatmap signals would overlap, and isolating the individual keypoints from the combined heatmap signals would be impractical.

To address these issues surrounding the use of heatmaps, we propose to adapt a deep learning-based object detector to perform keypoint detection by modeling keypoints as objects. To this end, we introduce the notion of a \textit{keypoint bounding box}, which is a small bounding box that represents a keypoint location using its center. During the training phase, the keypoint detection network is optimized in the same manner as an object detector, i.e., using a loss function based on the intersection over union (IoU) of the predicted and target keypoint bounding boxes. However, at inference time, the predicted keypoints are taken as the centers of the predicted keypoint bounding boxes. Notably, our keypoint detection method may be applied to any task that requires detecting an unknown number of keypoints, where there may be multiple instances of the same keypoint class in the input image.

\begin{figure}
\vspace{-12pt}
\centering
    \includegraphics[width=\linewidth]{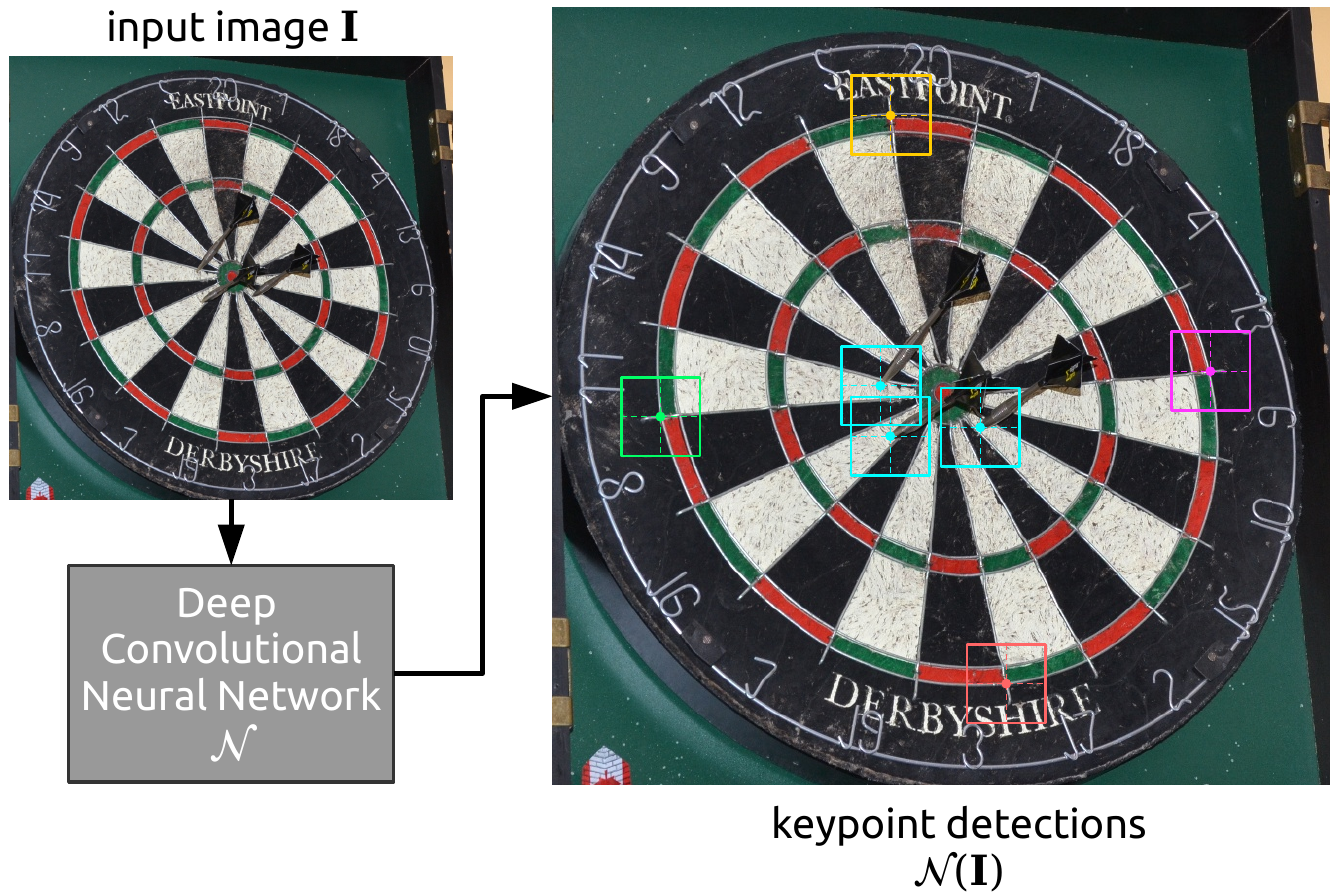}
\caption{Example inference of the deep neural network $\mathcal{N}$ that models keypoints as objects in order to map an input image $\mathbf{I}$ to the dart coordinates $\mathbf{\hat{P}_d}$ and four calibration points $\mathbf{\hat{P}_c}$. Here, the keypoint bounding box size is 10\% of the image size.}
\vspace{-12pt}
\label{fig:cnn}
\end{figure}

To apply the proposed keypoint detection method to predict dart scores, we develop a deep convolutional neural network $\mathcal{N}(\cdot)$, which takes as input an RGB image $\mathbf{I}\in\mathbb{R}^{h\times w\times 3}$ and outputs the locations of four dartboard calibration points $\mathbf{\hat{P}_c} = \{(\hat{x}_i, \hat{y}_i)\}_{i=1}^4$ and $D$ dart landing positions $\mathbf{\hat{P}_d} = \{(\hat{x}_j, \hat{y}_j)\}_{j=1}^D$ in the image coordinates, i.e., $\{(\hat{x}, \hat{y})\in\mathbb{R}^2\colon0 < \hat{x} < w, 0 < \hat{y} < h\}$:
\vspace{-4pt}
\begin{equation}
    \mathcal{N}(\mathbf{I}) = (\mathbf{\hat{P}_c}, \mathbf{\hat{P}_d}).
    \vspace{-4pt}
\end{equation}
The function of the network $\mathcal{N}$ is illustrated in Fig.\ \ref{fig:cnn}. During training, we model the four calibration points as separate classes and the dart locations as a fifth class. We utilize the state-of-the-art YOLOv4~\cite{bochkovskiy2020yolov4} as the base network for its superior computational efficiency and accuracy in the object detection task. More specifically, we implement its lightweight version, YOLOv4-tiny~\cite{wang2020scaled}, to help support potential mobile deployment. We conduct several ablation experiments to investigate the influence of the keypoint bounding box size on keypoint localization, and investigate the benefit of multiple data augmentation strategies that were developed specifically for the task at hand.

\subsection{Dart Score Prediction}
\vspace{-4pt}
The four chosen calibration points are located on the outer edge of the double ring, at the intersections of 5 and 20, 13 and 6, 17 and 3, and 8 and 11 (indicated in Fig.\ \ref{fig:cnn}). Using the correspondence between the detected set of calibration points $\mathbf{\hat{P}_c}$ and their known locations on the dartboard, the estimated homography matrix $\hat{H}$, which is a 3x3 invertible matrix that transforms a point in the image plane to a corresponding point in the dartboard plane, has a closed-form solution and is computed via a direct linear transform algorithm~\cite{Hartley2004}. To obtain the corresponding points $\mathbf{\hat{P}_c'}$ and $\mathbf{\hat{P}_d'}$ in the dartboard plane, the transformation is performed as follows
\vspace{-4pt}
\begin{equation}
\begin{bmatrix}\hat{x}'\cdot\lambda \\ \hat{y}'\cdot\lambda \\ \lambda\end{bmatrix} = \hat{H}
\begin{bmatrix}\hat{x} \\ \hat{y} \\ 1\end{bmatrix}
\vspace{-4pt}
\end{equation}
where $x'$ and $y'$ are the predicted coordinates of a point in the dartboard plane. The center of the dartboard is computed as the mean of the transformed calibration points $\mathbf{\hat{P}_c'}$. The radius of the outer edge of the double ring is computed as the mean of the distances between $\mathbf{\hat{P}_c'}$ and the center. Knowing the ratios between the radii of all circles in the scoring area\footnote{Dartboard specification referenced from the British Darts Organisation Playing Rules: \href{https://www.bdodarts.com/images/bdo-content/doc-lib/B/bdo-playing-rules.pdf}{https://www.bdodarts.com/images/bdo-content/doc-lib/B/bdo-playing-rules.pdf}}, the dart score predictions $\mathbf{\hat{S}}$ are obtained by classifying the points $\mathbf{\hat{P}_d'}$ into the dartboard sections based on their distance from the center and their angle from a reference direction, i.e., using polar coordinates. We refer to the represented mapping of dartboard image keypoints to dart scores as the scoring function $\phi(\cdot)$:
\vspace{-4pt}
\begin{equation}
\mathbf{\hat{S}} = \phi(\mathbf{\hat{P}_c}, \mathbf{\hat{P}_d}).
\end{equation}


\subsection{Data Augmentation for Dart Score Prediction}
\label{sec:aug}
\vspace{-4pt}

To help regularize the training of $\mathcal{N}$, which in turn improves the accuracy of the dart score predictions, we propose several task-specific data augmentation strategies. Some of the strategies change the positions of the darts while keeping the calibration points fixed, so as to not confuse the network regarding to the relative positioning of the calibration points, while others change the positions of all the keypoints. Each augmentation strategy is described below. For \textit{dartboard flipping} and \textit{dartboard rotation}, the augmentation is performed on the transformed $\mathbf{I}'$, $\mathbf{P_c}'$, and $\mathbf{P_d}'$, before transforming back to the original perspective using the inverse homography matrix $H^{-1}$.

\smallskip\noindent\textbf{Dartboard Flipping.} $\mathbf{I}'$ and $\mathbf{P_d}'$ are randomly flipped horizontally and/or vertically while $\mathbf{P_c}'$ remains fixed.

\begin{figure*}
\vspace{-6pt}
  \begin{subfigure}[b]{0.245\textwidth}
    \includegraphics[width=\linewidth]{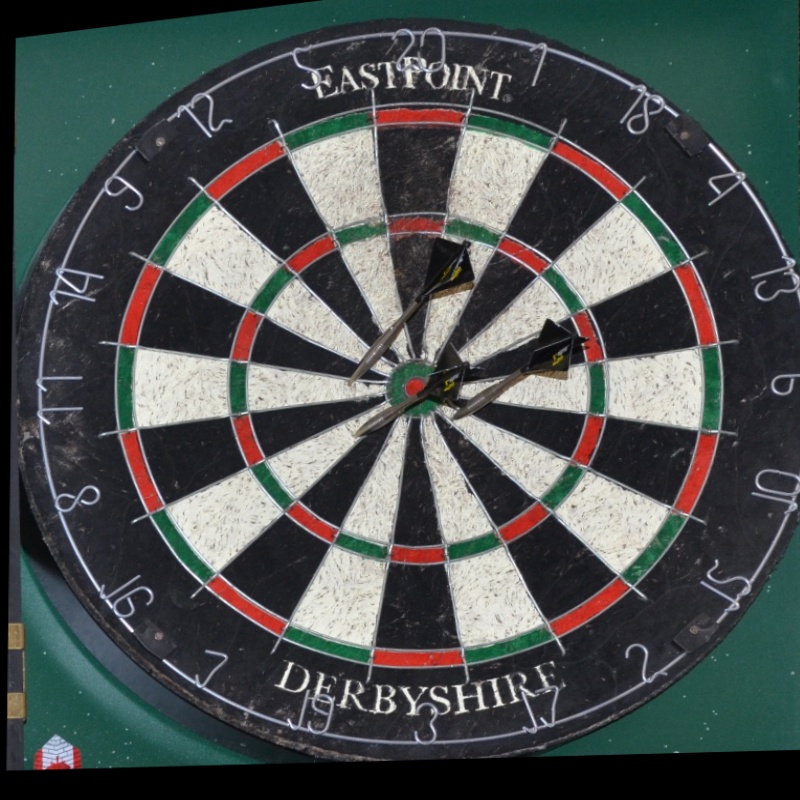}
    \caption{$s_\rho = 0$ (warped perspective)}
  \end{subfigure}
  \begin{subfigure}[b]{0.245\textwidth}
    \includegraphics[width=\linewidth]{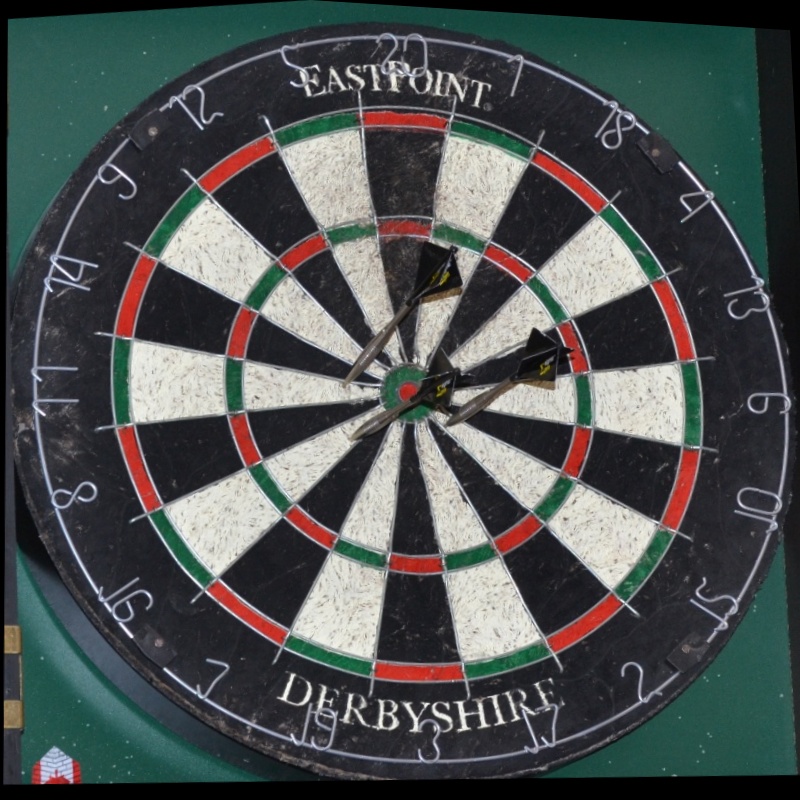}
    \caption{$s_\rho = 0.5$ (warped perspective)}
  \end{subfigure}
  \begin{subfigure}[b]{0.245\textwidth}
    \includegraphics[width=\linewidth]{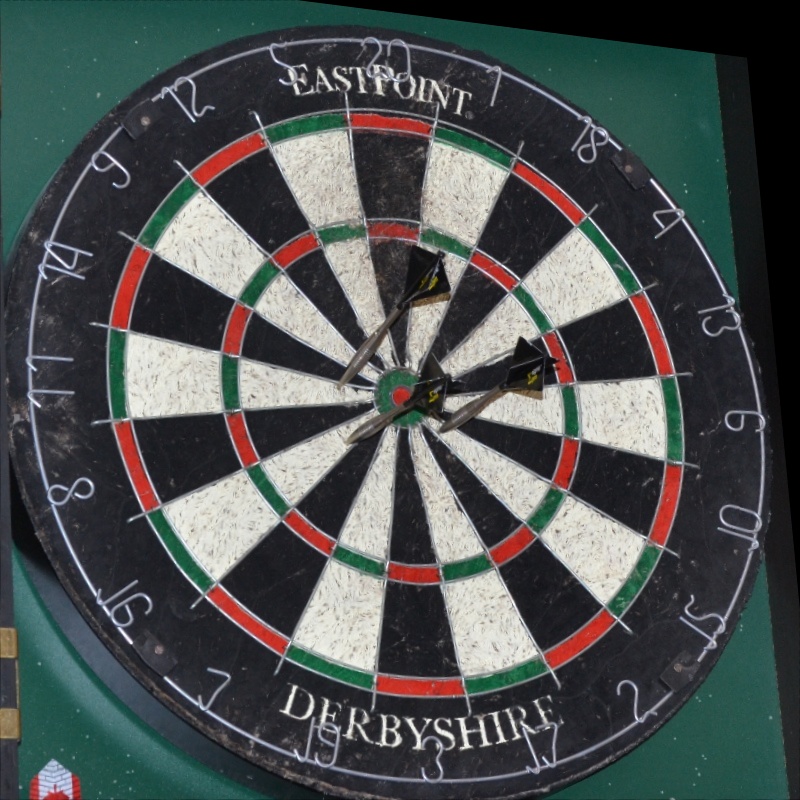}
    \caption{$s_\rho = 1$ (original perspective)}
  \end{subfigure}
  \begin{subfigure}[b]{0.245\textwidth}
    \includegraphics[width=\linewidth]{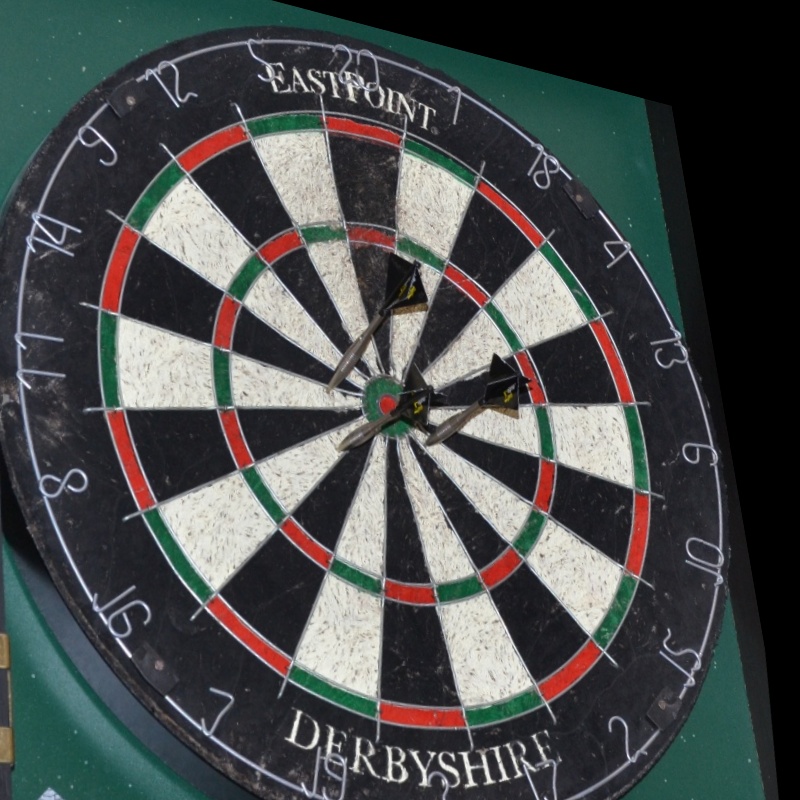}
    \caption{$s_\rho = 2$ (warped perspective)}
  \end{subfigure}
\vspace{-4pt}
\caption{Demonstrating the effect of the perspective warping data augmentation strategy. The depicted images are examples of when the non-diagonal elements of the inverse homography $H^{-1}$ matrix are scaled equally by $s_\rho\in\{0, 0.5, 1, 2\}$. During training, the non-diagonal elements of $H^{-1}$ are scaled randomly and separately.} 
\vspace{-12pt}
\label{fig:setup}
\end{figure*}

\smallskip\noindent\textbf{Dartboard Rotation.} $\mathbf{I}'$ and $\mathbf{P_d}'$ are randomly rotated (in the image plane) in the range [-\ang{180}, \ang{180}] using a step size of \ang{18} or \ang{36} while $\mathbf{P_c}'$ remains fixed. A step size of \ang{18} degrees keeps the dartboard sections aligned, where either a white or black section may appear at the top. A step size of \ang{36} ensures only black sections appear at the top.

\smallskip\noindent\textbf{Small Rotations.} To account for dartboards that are not perfectly vertically aligned, we apply small random rotations to $\mathbf{I}$, $\mathbf{P_c}$, and $\mathbf{P_d}$ in the range [-\ang{2}, \ang{2}].

\smallskip\noindent\textbf{Perspective Warping.} To help generalize to various camera angles, we randomly warp the perspective of the dartboard images. To implement perspective warping in a principled manner, $H^{-1}$ is randomly perturbed before $\mathbf{I}'$, $\mathbf{P_c}'$, and $\mathbf{P_d}'$ are transformed back to the original perspective. We introduce a hyperparameter $\rho$ to control the amount of perspective warping. Specifically, the non-diagonal elements of $H^{-1}$ are randomly scaled by factors sampled from a uniform distribution in the range [0, $\rho$]. 

\begin{figure*}
\vspace{-12pt}
  \begin{subfigure}[b]{0.28\textwidth}
    \centering
    \includegraphics[width=\linewidth]{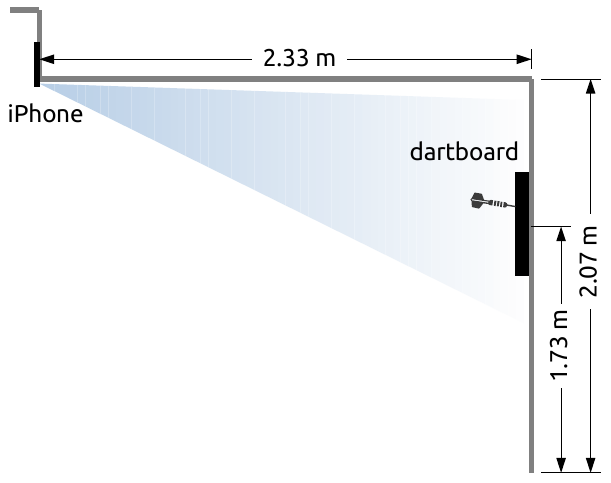}
    \caption{$\mathcal{D}_1$ data collection setup} \label{fig:setup-a}
  \end{subfigure}
  \hspace*{\fill}   
  \begin{subfigure}[b]{0.23\textwidth}
    \centering
    \includegraphics[width=\linewidth, angle=90]{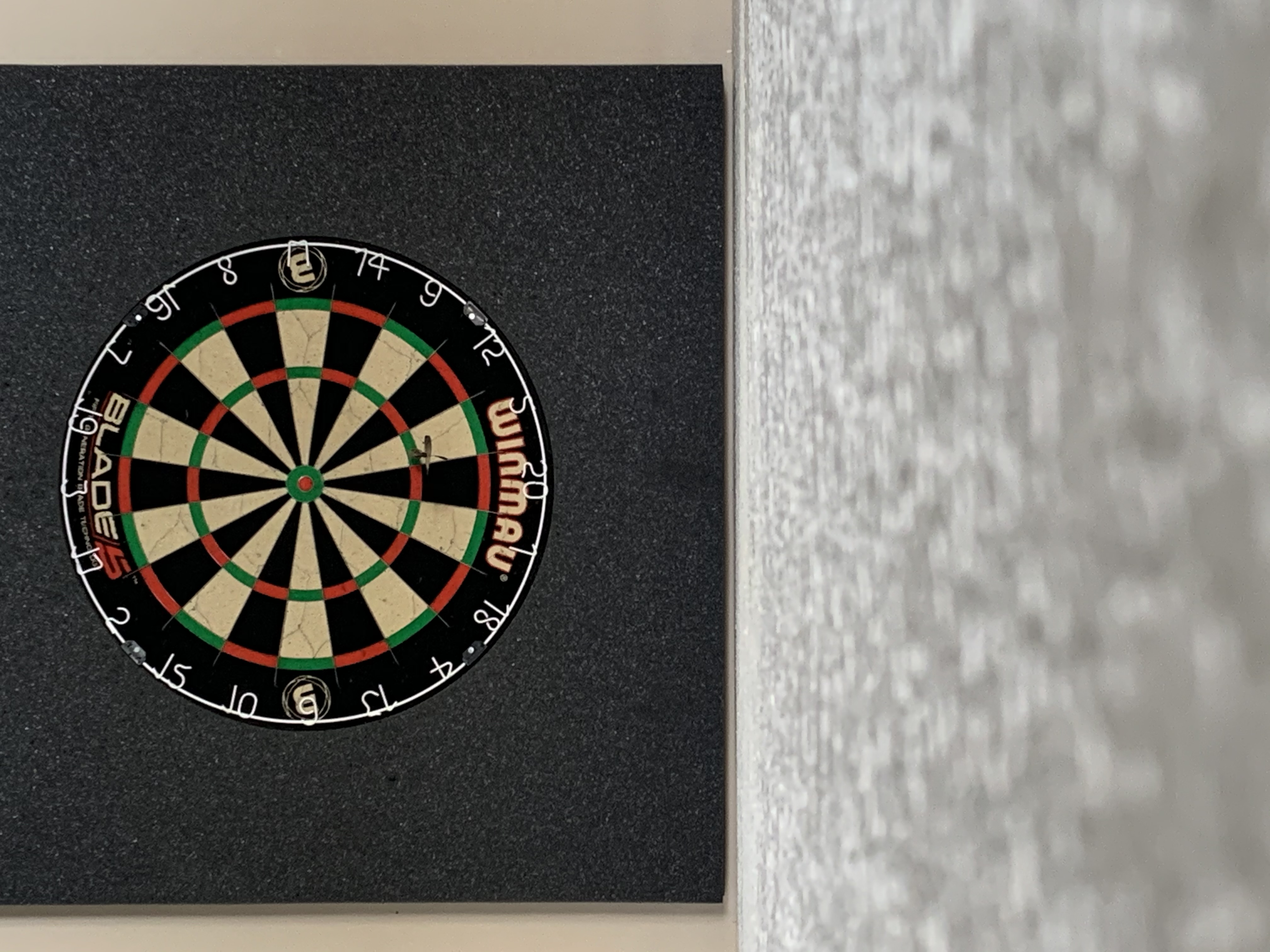}
    \caption{{$\mathcal{D}_1$ sample image}} \label{fig:setup-b}
  \end{subfigure}
  \hspace*{\fill}
  \begin{subfigure}[b]{0.23\textwidth}
    \centering
    \includegraphics[width=\linewidth, angle=-90]{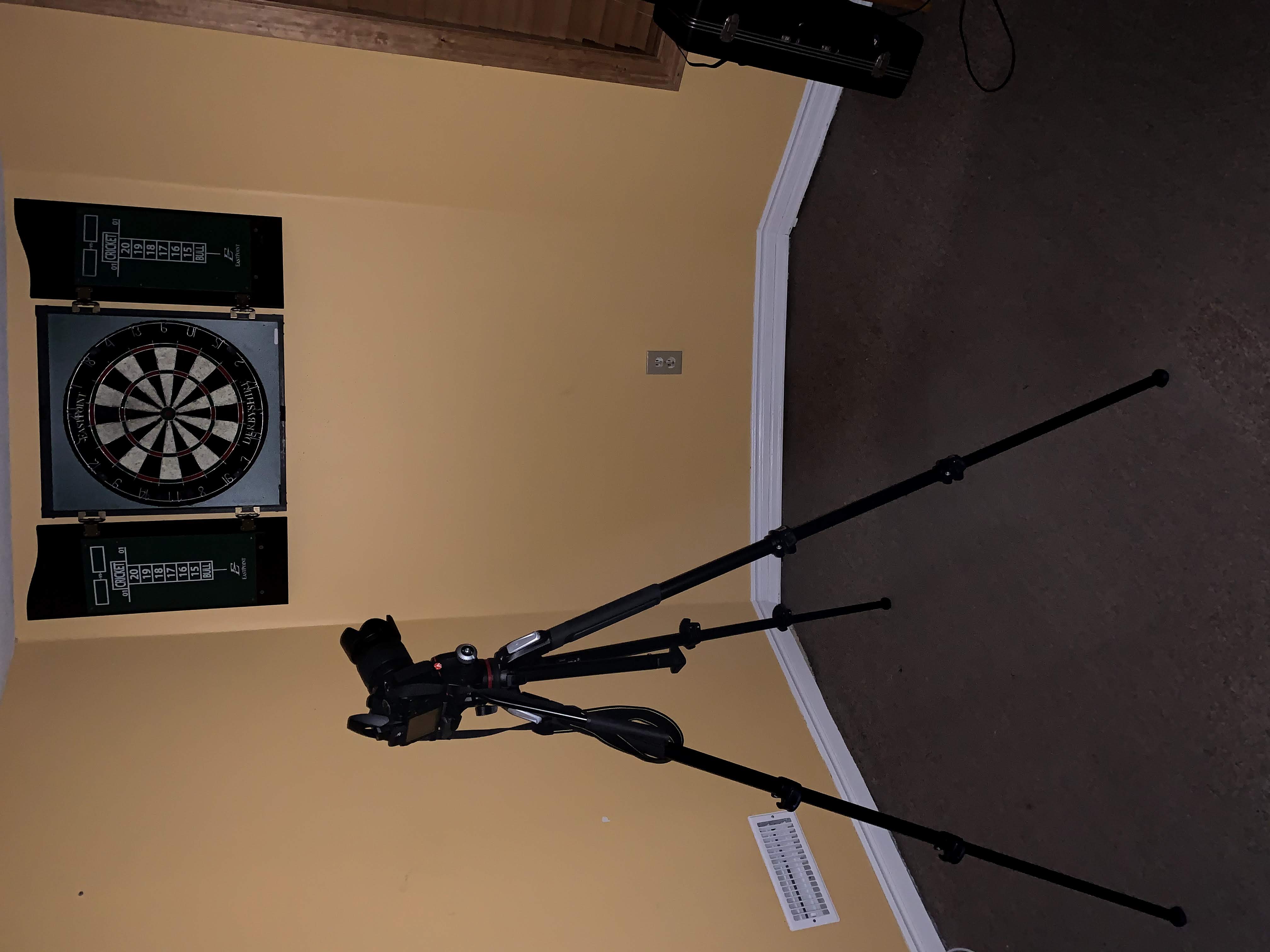}
    \caption{$\mathcal{D}_2$ data collection setup} \label{fig:setup-c}
  \end{subfigure}
  \hspace*{\fill}
  \begin{subfigure}[b]{0.23\textwidth}
    \centering
    \includegraphics[width=\linewidth]{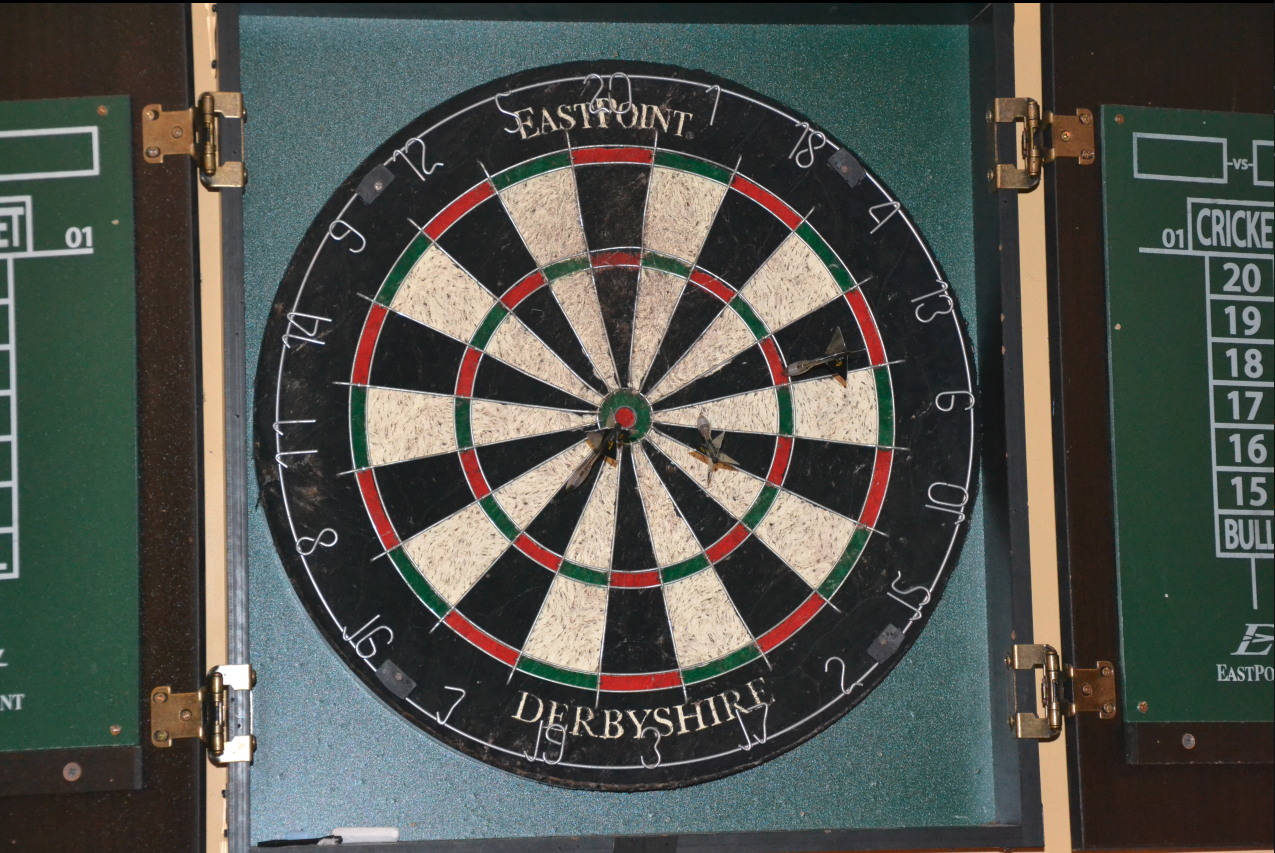}
    \caption{$\mathcal{D}_2$ sample image} \label{fig:setup-d}
  \end{subfigure}
\caption{Data collection setups and sample images.} 
\vspace{-12pt}
\label{fig:warping}
\end{figure*}

To illustrate the effect of the augmentation, it is helpful to consider a scenario when the non-diagonal elements of $H^{-1}$ are scaled equally by $s_\rho$. When $s_\rho=0$, $H^{-1}$ approximately equals the identity matrix and the image remains in a face-on perspective, i.e., with a perfectly circular dartboard. When $s_\rho=1$, $H^{-1}$ is unchanged and the image is transformed back to its original perspective. For $0 < s_\rho < 1$, the warped perspective is effectively an interpolation between the face-on and original perspective. When $s_\rho > 1$, the warped perspective is effectively an extrapolation of the original perspective. Example warped images for $s_\rho\in\{0, 0.5, 1, 2\}$ are shown in Fig.\ \ref{fig:warping}. During training, the non-diagonal elements of $H^{-1}$ are scaled separately and randomly to increase variation.

\section{Datasets}
\vspace{-4pt}
A total of 16,050 dartboard images containing 32,027 darts were manually collected and annotated. The images originate from two different dartboard setups, and thus were separated into two datasets $\mathcal{D}_1$ and $\mathcal{D}_2$. The primary dataset $\mathcal{D}_1$ includes 15k images collected using a smartphone camera positioned to capture a face-on view of the dartboard. The second dataset $\mathcal{D}_2$ contains the remaining 1050 images, which were taken from various camera angles using a digital single-lens reflex (DSLR) camera mounted on a tripod. 

\subsection{Image Data Collection: Dataset $\mathbf{\mathcal{D}_1}$}
\vspace{-4pt}
A Winmau Blade 5 dartboard was mounted at regulation height (1.73m) in a section of a room with a lowered ceiling. The transition wall between the low and high ceilings ran parallel with the dartboard, which allowed an iPhone XR to be mounted overhead near the throw line. A piece of double-sided adhesive tape was placed on the rear side of the iPhone plastic case, and the case was secured to the transition wall in-line with the center of the dartboard, and in an upside-down position such that the iPhone camera extended just beyond the low ceiling. A diagram of the data collection setup is shown in Fig.\ \ref{fig:setup-a}, and a sample image from the iPhone XR is shown in Fig.\ \ref{fig:setup-b}. 


Steel-tip darts were thrown by left- and right-handed beginner and intermediate players. An image was captured after each throw, and the darts were retrieved after three darts were thrown. Capturing an image after each throw increased the number of unique images in the dataset and also helped during annotation to identify the landing position of a dart that was occluded by a subsequently thrown dart. Moreover, a thrown dart would occasionally impact a dart already on the board, changing the orientation of the previously thrown dart and thus its appearance from the previous image. A variety of different games (e.g., 501, Cricket, Around the World, etc.) were played to distribute the data across all sections of the dartboard.


Several windows were in the vicinity of the dartboard, and images were collected during the day and at night, which provided a variety of natural and artificial lighting conditions. In some lighting conditions, the darts cast shadows on the dartboard. A number of edge cases were encountered during the data collection. For example, flights would occasionally dislodge upon striking the dartboard and fall to the ground. In rare cases, the tip of a thrown dart would penetrate the stem of a previously thrown dart and reside there, never reaching the dartboard. The images were collected over 36 sessions. In four sessions amounting to 1,200 images, the score of each dart was also recorded. This information was used to assess the accuracy of the annotation process (see Section \ref{sec:annotation} for details). The dataset was split at the session level to generate train, validation, and test subsets containing 12k, 1k, and 2k images, respectively.

\subsection{Image Data Collection: Dataset $\mathbf{\mathcal{D}_2}$}
\vspace{-4pt}
The dartboard setup used in this dataset included an EastPoint Derbyshire Dartboard and Cabinet Set. Images were collected using a Nikon D3100 DSLR camera mounted on a tripod. The data collection setup is depicted in Fig.\ \ref{fig:setup-c}, and a sample image is shown in Fig.\ \ref{fig:setup-d}. Similar to dataset $\mathcal{D}_1$, images were taken after each thrown dart. A total of 1,050 images were collected over 15 sessions, and the camera was repositioned in each session to capture a variety of camera angles. The dataset also includes a variety of lighting conditions resulting from various combinations of natural and artificial light (from ceiling lights as well as the camera flash). The dataset was split at the session level to generate train, validation, and test subsets containing 830, 70, and 150 images, respectively.

While the images in this dataset may not be consistent with those encountered during actual deployment on an edge device, the primary purpose of this dataset is to test the automatic dart scoring system in a more challenging scenario including various camera angles and limited training data. We also use this dataset to investigate whether the knowledge learned from one dartboard setup can be transferred to another. 

\subsection{Keypoint Annotation}
\label{sec:annotation}
\vspace{-4pt}
All images were annotated by a single person using a custom-made annotation tool developed in Python 3. For each image, up to seven keypoints ($x$, $y$) were identified, including the four dartboard calibration points $\mathbf{P_c}$, and up to three dart landing positions $\mathbf{P_d}$. In face-on views of the dartboard, the exact position of a dart was often not visible due to self-occlusion, as the dart barrel and flight tended to obstruct the view of the dart tip. Occasionally, there was occlusion from other darts as well. In such cases, the dart landing position was inferred at the discretion of the annotator. To assess the accuracy of the labeling process, the scores of the labeled darts were computed using the scoring function $\phi(\mathbf{P_c},\mathbf{P_d})$ and were compared against the actual scores of the 1,200 darts that were recorded during the collection of the $\mathcal{D}_1$ images. The labeled and actual scores matched for 97.6\% of the darts. The dataset annotations also include square bounding boxes that enclose the dartboard. These were automatically generated using $\mathbf{P_c}$.

\begin{figure}
\vspace{-8pt}
\centering
    \includegraphics[trim={12mm, 15mm, 12mm, 12mm}, clip, width=0.5\linewidth]{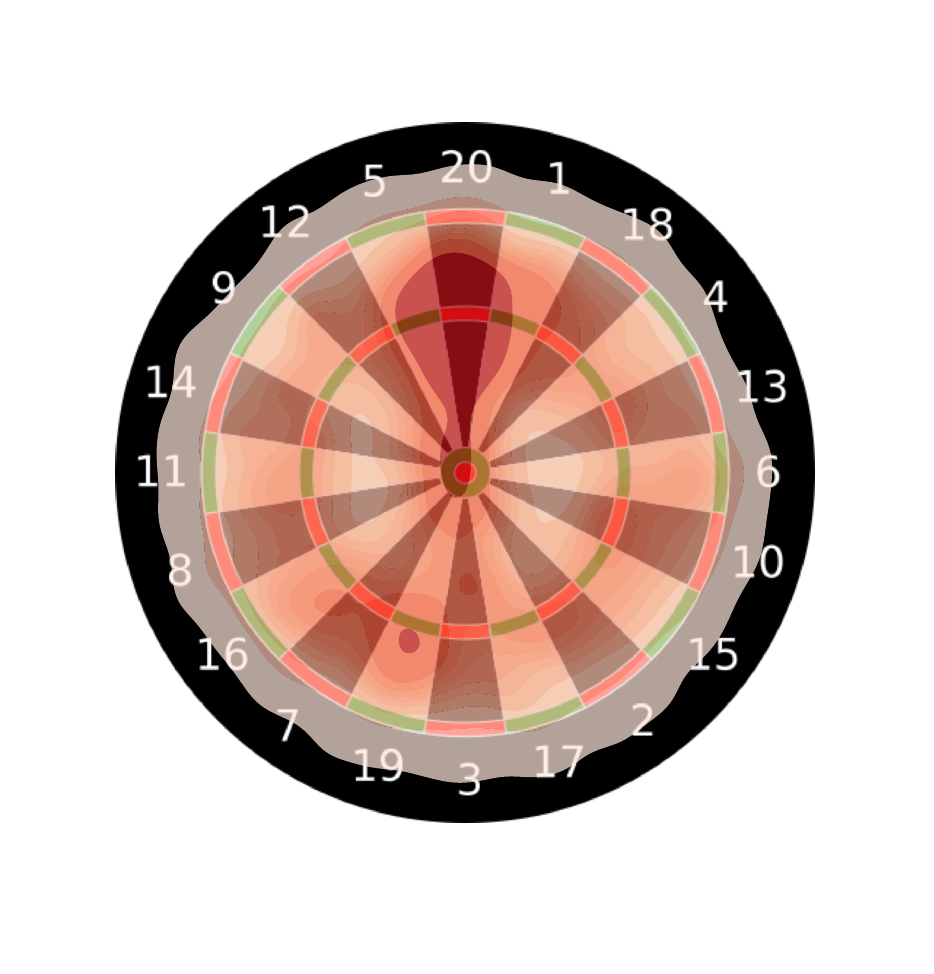}
\caption{Probability distribution of the labelled dart positions in $\mathcal{D}_1$ and $\mathcal{D}_2$ (32,027 darts in total). The darker regions represent higher frequency landing positions.}
\vspace{-12pt}
\label{fig:db-heatmap}
\end{figure}

\begin{figure}
\centering
\includegraphics[trim={0, 2mm, 0, 0}, clip, width=0.9\linewidth]{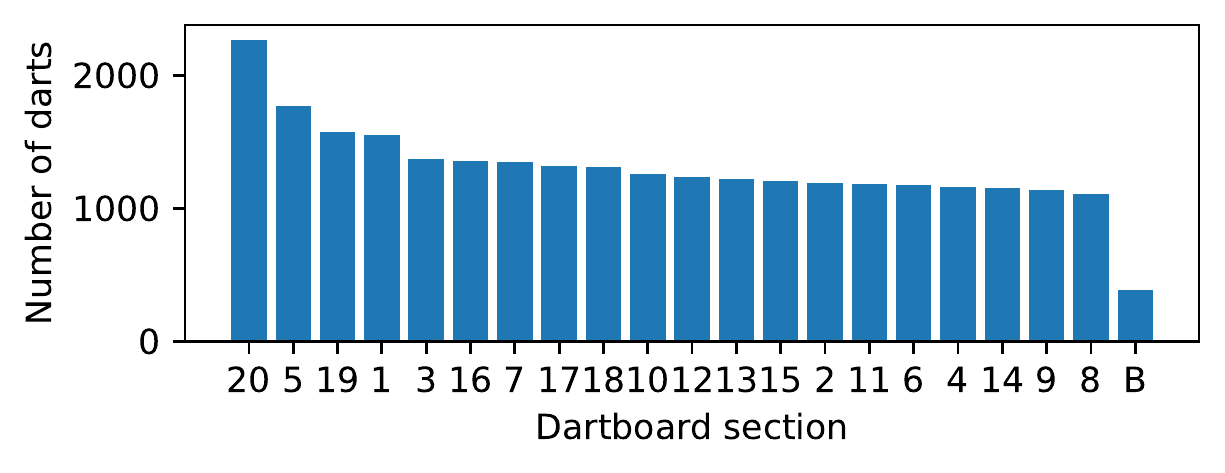}
\vspace{-10pt}
\caption{Number of labelled darts in each section of the dartboard. `B' includes bull and double bull.}
\vspace{-12pt}
\label{fig:dart-counts}
\end{figure}

The probability distribution of the labelled dart positions is plotted in Fig.\ \ref{fig:db-heatmap}, and shows that T20 and T19 were among the most highly targeted areas of the dartboard (darker regions). The dart counts per scoring section are provided in Fig.\ \ref{fig:dart-counts}, and show a fairly uniform distribution, with the exception of 20, 19, and their neighbouring sections, which were more frequent, and the bullseye, which was less frequent. 

\subsection{Percent Correct Score}
\vspace{-4pt}
A meaningful accuracy metric for detection should take into account false positives as well as false negatives. The mean average precision (mAP) is a common accuracy metric used in object detection that takes into account false positives and negatives by averaging the area under the precision-recall curve at various IoU thresholds. However, mAP can be difficult to interpret and put into context. We therefore introduce a task-specific accuracy metric that is easy to interpret and takes into account false positives and false negatives through evaluation of the total score of the dartboard, as opposed to the individual dart scores. We refer to this metric as the \textit{Percent Correct Score} (PCS), and it represents the percentage of dartboard image samples whose predicted total score $\sum\mathbf{\hat{S}}$ matches the labeled total score $\sum\mathbf{S}$. More explicitly, over a dataset with $N$ images, the PCS is computed as follows:
\vspace{-4pt}
\begin{equation}
    PCS = \frac{100}{N} \sum^N_{i=1} \delta \left(\left(\sum \mathbf{\hat{S}}_i - \sum\mathbf{S}_i\right) = 0\right)\%
\end{equation}

\section{Experiments}
\vspace{-4pt}
This section contains results from various ablation experiments investigating the influence of different training configurations on the validation accuracy of DeepDarts. Following the ablation experiments, optimal training configurations for $\mathcal{D}_1$ and $\mathcal{D}_2$ are proposed and the final test results are reported. 

\subsection{Implementation Details}
\vspace{-4pt}
The ground-truth dartboard bounding boxes were used to extract a square crop of the dartboard from the raw images. The cropped dartboard images were then resized to the desired input size. In an actual application of DeepDarts, we argue that the user could manually draw a bounding box around the dartboard in the camera view, and additional scaling / translation augmentation could be used to account for the variability in the drawn bounding box. Alternatively, $\mathcal{N}$ could be trained to simultaneously detect the dartboard and keypoints from the raw images, but doing so could potentially have a negative effect on the accuracy of the dart score predictions. 

In all experiments, YOLOv4-tiny~\cite{wang2020scaled} was used as the base network for keypoint detection. All networks were trained using the Adam optimizer~\cite{kingma2014adam} with a cosine decay learning rate schedule and an initial learning rate of 0.001~\cite{loshchilov2016sgdr}. The loss function used was the same as in the original YOLOv4 implementation~\cite{bochkovskiy2020yolov4, wang2020scaled}, and is based on CIoU~\cite{zheng2020distance}. 

During inference, the predicted keypoint bounding boxes were filtered using an IoU threshold of 0.3 and a confidence threshold of 0.25. If extra calibration points were detected, the points with the highest confidence were used. If one calibration point was missed, its position was estimated based on the positions of the three detected calibration points. If two or more calibration points were missed, the sample was assigned a total score of 0. Further implementation details are provided in the experiment descriptions as needed.

\subsection{Ablation Experiments}
\vspace{-4pt}
\noindent\textbf{Data Augmentation.} In addition to the data augmentation strategies proposed in Section~\ref{sec:aug}, we also investigate the benefits of small translations (jitter). Each data augmentation strategy was tested individually and was applied with a probability of 0.5 over 20 epochs of training. An input size of 480 was used and the keypoint bounding box size was set to 12 px. The $\mathcal{D}_1$ experiments were run on four GPUs using a batch size of 32 per GPU. The $\mathcal{D}_2$ experiments were run on two GPUs using a batch size of 4 per GPU, resulting in approximately the same number of training iterations as the $\mathcal{D}_1$ experiments. The validation PCS is reported for each data augmentation strategy in Table~\ref{tab:aug}. Due to the limited number of samples in the $\mathcal{D}_2$ validation set, there was significant variability in the PCS from one run to the next. We therefore report the mean PCS over five runs for the $\mathcal{D}_2$ experiments. 

\begin{table}[]
    \centering
    \begin{tabular}{l|c|c}
        \hline
        Augmentation Strategy & \makecell{$\mathcal{D}_1$ \texttt{val} PCS} & \makecell{$\mathcal{D}_2$ \texttt{val} PCS}\\
        \hline
        None & 77.5 & 57.7 \\
        Dartboard Flipping & 83.1 (+5.6) & 60.6 (+2.9) \\
        Dartboard Rot. (\ang{18}) & 83.2 (+5.7) & 58.3 (+0.6) \\
        Dartboard Rot. (\ang{36}) & 84.3 (+6.8) & 60.0 (+2.3) \\
        Small Rotations & 82.1 (+4.6) & 62.6 (+4.9) \\
        Perspective Warping & 80.3 (+2.8) & 64.9 (+7.1) \\ 
        Jitter & 81.7 (+4.2) & 63.7 (+6.0) \\
        \hline
    \end{tabular}
    \vspace{-8pt}
    \caption{Effect of the proposed data augmentation strategies. $\mathcal{D}_2$ PCS averaged over five runs.}
    \vspace{-8pt}
    \label{tab:aug}
\end{table}

All of the data augmentation strategies improved the validation PCS. On $\mathcal{D}_1$, dartboard rotations with a step size of \ang{36} provided the greatest benefit, and an improvement of 1.1 PCS over step sizes of \ang{18}, suggesting that the placement of the section colours had an effect on learning. Perspective warping was the most effective on $\mathcal{D}_2$ as it helped generalize to the various camera angles in the dataset, and provided an improvement of 7.1 PCS. 

\begin{figure}
\centering
\includegraphics[trim={0, 2mm, 0, 0}, clip, width=1\linewidth]{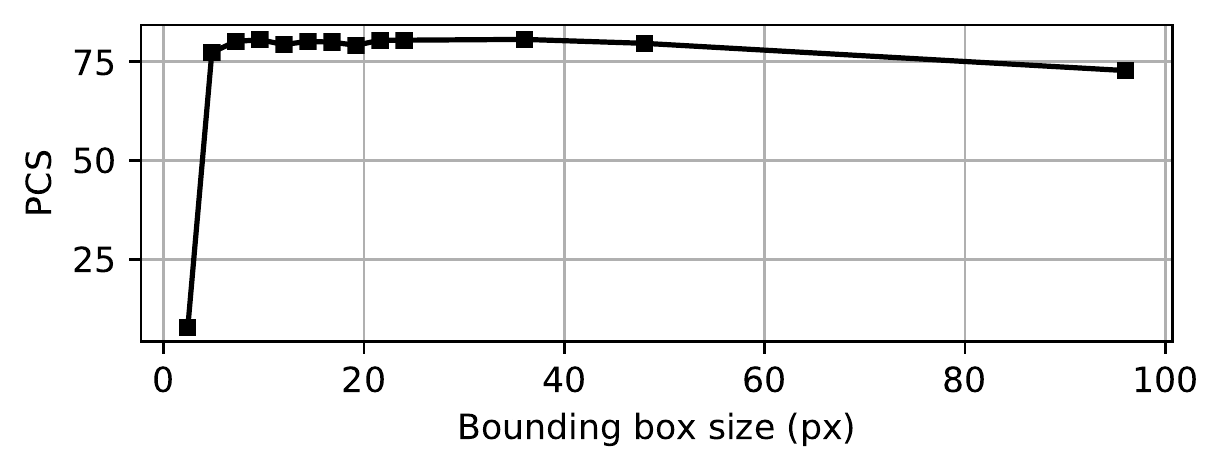}
\caption{Influence of keypoint bounding box size on $\mathcal{D}_1$ validation PCS using a fixed input size of 480.}
\label{fig:bbox}
\vspace{-6pt}
\end{figure}

\smallskip\noindent\textbf{Keypoint Bounding Box Size.} The keypoint bounding box size was varied from 0.5\% (2.4 px) to 20\% (96 px) of the input size, which was fixed at 480. No data augmentation was used and each network was trained on $\mathcal{D}_1$ for 20 epochs using two GPUs and a batch size of 32 per GPU. The validation accuracies are plotted in Fig.\ \ref{fig:bbox}. It is evident from the results that using very small keypoint bounding boxes, in this case less than 1\% or 4.8 px, is detrimental to training. Above this threshold, however, the accuracy is not very sensitive to the keypoint bounding box size, but begins to decrease slightly above a relative keypoint bounding box size of 7.5\% (36 px). 

\begin{figure}
\vspace{-10pt}
\centering
\includegraphics[trim={0, 0, 0, 0}, clip, width=0.95\linewidth]{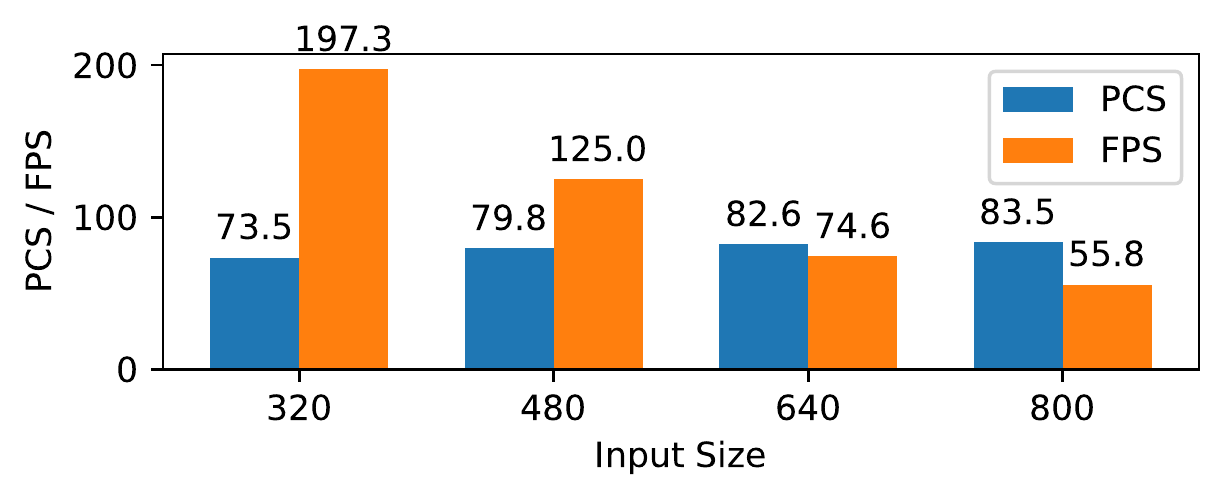}
\vspace{-8pt}
\caption{Influence of input size on $\mathcal{D}_1$ validation PCS and inference speed.}
\vspace{-4pt}
\label{fig:input-size}
\end{figure}

\smallskip\noindent\textbf{Input Size.} To investigate the trade-off between accuracy and inference speed, the input size was varied from 320 to 800. The keypoint bounding box sizes were set to 2.5\% of the input size, and each network was trained for 20 epochs on $\mathcal{D}_1$ using two GPUs. For input sizes of 640 and 800, the batch sizes were reduced to 24 and 16 per GPU, respectively, to accommodate limited GPU memory. The validation accuracies are provided in Fig.\ \ref{fig:input-size}, and show diminishing returns for a linear increase in the input resolution. The inference speed, measured in frames per second (FPS) using a batch size of 1, was inversely correlated with the input size, but the system still achieved real-time speeds greater than 30 FPS using the maximum input size of 800. 

\smallskip\noindent\textbf{Transfer Learning.} To investigate whether the knowledge learned from one dartboard can be transferred to another, $\mathcal{N}$ was initialized with the weights learned on $\mathcal{D}_1$ and then trained on $\mathcal{D}_2$. The effect of initializing with ImageNet~\cite{deng2009imagenet} weights was also tested. These experiments were run on two GPUs and the hyperparameters were consistent with those in the base case of the data augmentation experiments. The results are reported in Table \ref{tab:pretrain}. ImageNet pretraining improved the PCS on both datasets despite a marked dissimilarity between the two tasks. When transferring the $\mathcal{D}_1$ weights, it was found that the weights from all but the final convolutional layer should be transferred for effective training. The $\mathcal{D}_1$ weights provided an improvement of 10.0 PCS on $\mathcal{D}_2$, indicating a successful transfer of knowledge between the two independent dartboard setups. 

\begin{table}[]
    \centering
    \begin{tabular}{l|c|c}
        \hline
        Pretraining & $\mathcal{D}_1$ \texttt{val} PCS & $\mathcal{D}_2$ \texttt{val} PCS \\
        \hline
        None & 79.8 & 57.7 \\
        ImageNet & 82.2 (+2.4) & 61.7 (+4.0) \\
        Dataset $\mathcal{D}_1$ & -- & 67.7 (+10.0) \\
        \hline
    \end{tabular}
    \vspace{-8pt}
    \caption{Influence of transfer learning from ImageNet and dataset $\mathcal{D}_1$ on validation PCS. $\mathcal{D}_2$ PCS averaged over five runs.}
    \vspace{-12pt}
    \label{tab:pretrain}
\end{table}

\subsection{Final Training Configurations and Test Results}
\vspace{-4pt}
To maximize accuracy, $\mathcal{N}$ was trained for 100 epochs on each dataset using an input size of 800, a keypoint bounding box size of 2.5\%, and a combination of data augmentations. The overall probability of data augmentation was set to 0.8, after which each data augmentation strategy was applied at a rate of 0.5. For dartboard rotation, a step size of \ang{36} was used. On $\mathcal{D}_1$, perspective warping was omitted, and the network was initialized with the ImageNet weights. On $\mathcal{D}_2$, the network was initialized with the pretrained weights from $\mathcal{D}_1$, and we report the results for the model with the best test PCS over five runs. The experiments were run on two GPUs using batch sizes of 16 and 4 per GPU for datasets $\mathcal{D}_1$ and $\mathcal{D}_2$, respectively. The validation and test PCS are provided in Table \ref{tab:test}. The test PCS of DeepDarts was 94.7\% on $\mathcal{D}_1$, and 84.0\% on $\mathcal{D}_2$.

\smallskip\noindent\textbf{Failure Cases.} The most common failure mode was missed dart detections due to occlusion from other darts. In actual deployment, some of these errors could be accounted for as they would be detectable when a previous dart prediction with high confidence suddenly disappears. The second most common error occurred when darts were on the edge of a section and were incorrectly scored. In rare cases, the ground-truth labels were incorrect, darts were missed due to unusual dart orientations, or calibration points were missed due to dart occlusion. In future work, we recommend training the network to detect redundant calibration points to improve the accuracy of the system. To provide a sense of the error distribution, of the 24 errors on the $\mathcal{D}_2$ test set, two errors were caused by occluded calibration points, three errors were caused by incorrect scoring, and the remaining errors were undetected darts due to occlusion. 

\begin{table}[]
    \centering
    \begin{tabular}{l|c|c|c|c}
        \hline
        Method & \makecell{$\mathcal{D}_1$\\\texttt{val}\\PCS} & \makecell{$\mathcal{D}_1$\\\texttt{test}\\PCS} & \makecell{$\mathcal{D}_2$\\\texttt{val}\\PCS} & \makecell{$\mathcal{D}_2$\\\texttt{test}\\PCS}\\
        \hline
        DeepDarts & 92.4 & 94.7 & 87.1 & 84.0 \\
        \hline
    \end{tabular}
    \vspace{-8pt}
    \caption{Final validation and test PCS.}
    \label{tab:test}
    \vspace{-14pt}
\end{table}

\section{Conclusion}
\vspace{-4pt}
We introduce DeepDarts, a system for predicting dart scores from a single image taken from any camera angle. DeepDarts leverages a deep convolutional neural network to detect dartboard keypoints in a new deep learning-based approach to keypoint detection in which keypoints are modeled as objects. Our experiments demonstrate that our method can predict dart scores precisely and generalizes to various camera angles. In one dataset, the system predicted the correct total score in 94.7\% of the test images. In future work, DeepDarts should be trained on a larger dataset containing a greater variety of dartboard images to enable in the wild deployment.

\smallskip\noindent\textbf{Acknowledgements.} We acknowledge financial support from the Canada Research Chairs Program and the Natural Sciences and Engineering Research Council of Canada (NSERC). We also acknowledge the TensorFlow Research Cloud Program, an NVIDIA GPU Grant, and Compute Canada for hardware support. 

{\small
\bibliographystyle{ieee_fullname}
\bibliography{egbib}
}

\end{document}